\documentclass{article}
\usepackage{ijcai22}

\usepackage{times}
\usepackage{soul}
\usepackage{url}
\usepackage[hidelinks]{hyperref}
\usepackage[utf8]{inputenc}
\usepackage[small]{caption}
\usepackage{graphicx}
\usepackage{amsmath}
\usepackage{amsfonts,amssymb}
\usepackage{amsthm}
\usepackage{booktabs}
\usepackage{algorithm}
\usepackage{algorithmic}
\usepackage{makecell}
\usepackage{subfigure}

\title{An Improved Reinforcement Learning Algorithm for Learning to Branch}

\author{
Qingyu Qu$^{1,2}$
Xijun Li$^{3,2}$\footnote{Xijun Li is the corresponding author.}
Yunfan Zhou$^{2}$
Jia Zeng$^2$
Mingxuan Yuan$^2$
Jie Wang$^3$
Jinhu Lv$^1$
Kexin Liu$^1$\And
Kun Mao$^4$
\affiliations
$^1$Beihang University\\
$^2$Huawei Noah's Ark Lab\\
$^3$MIRA Lab, USTC\\
$^4$Huawei Cloud Co.
\emails
\{quqingyu, skxliu\}@buaa.edu.cn,
\{xijun.li, zhouyunfan, zeng.jia, Yuan.Mingxuan, maokun\}@huawei.com,
jhlu@iss.ac.cn,
jiewangx@ustc.edu.cn
}

\begin{document}

\maketitle

\begin{abstract}
    Most combinatorial optimization problems can be formulated as mixed integer linear programming (MILP), in which branch-and-bound (B\&B) is a general and widely used method.
    Recently, learning to branch has become a hot research topic in the intersection of machine learning and combinatorial optimization.
    In this paper, we propose a novel reinforcement learning-based B\&B algorithm.
    Similar to offline reinforcement learning, we initially train on the demonstration data to accelerate learning massively. With the improvement of the training effect, the agent starts to interact with the environment with its learned policy gradually.
    It is critical to improve the performance of the algorithm by determining the mixing ratio between demonstration and self-generated data. Thus, we propose a prioritized storage mechanism to control this ratio automatically.
    In order to improve the robustness of the training process, a superior network is additionally introduced based on Double DQN, which always serves as a Q-network with competitive performance.
    We evaluate the performance of the proposed algorithm over three public research benchmarks and compare it against strong baselines, including three classical heuristics and one state-of-the-art imitation learning-based branching algorithm.
    The results show that the proposed algorithm achieves the best performance among compared algorithms and possesses the potential to improve B\&B algorithm performance continuously.
\end{abstract}

\section{Introduction}

The mixed integer linear programming (MILP) is one of the most widely-used mathematical formulations in practical scenarios of optimization, such as planning and scheduling, bin packing, resource allocation, etc.
Generally, the vast majority of MILP problems are NP-hard, which makes it inevitable to compromise in terms of solving speed, precision, generalization, etc. when using traditional methods to solve them.
The NP-hardness is mainly due to the scale, such as the number of integer variable.
Therefore, some scholars pay attention to improving the solutions of MILP by adopting some machine learning algorithms such as the imitation learning, reinforcement learning, etc.~\cite{huang2021branch}.
The branch-and-bound (B\&B) algorithm, together with the various skills for increasing its efficiency, constitute one of the cores of MILP.
B\&B algorithm enumerates the candidate solutions systematically by means of state space search, in which the set of candidates solutions is considered to form a search tree with the full set at the root.
The efficiency of B\&B algorithm mainly depends on branching variable selection and node selection.
In this paper, we concentrate on the former.
Usually, choosing good variables to branch on can lead to a dramatic reduction in terms of the number of nodes needed to solve an instance~\cite{Alvarez14asupervised}.

At present, there is still no commonly accepted method for the strategy of branching variable selection. 
Traditional methods are mostly simple heuristic rules, such as the Most Infeasible branching, Pseudo-Cost branching (PC), Strong Branching (SB), Hybrid Strong/Pseudocost Branching, Pseudocost branching with strong branching initialization, Reliability Branching (RB), etc.~\cite{Achterberg:2}, in which PC and SB approaches are the two most common heuristic rules.
PC is a sophisticated rule in the sense that it keeps a history of the success of the variables on which already has been branched~\cite{Benichou:3}.
Despite tiny computation, PC relies on human intuition and extensive engineering, requiring significant manual tuning.
SB is to evaluate which of the fractional candidates gives the largest progress before actually branching on any of them~\cite{Applegate:4}.
This approach can lead to the smallest search tree currently known. However, it increases the computation significantly.

\cite{Alvarez14asupervised} adopted machine learning algorithms early to learn the strategies of branching variable selection in the B\&B algorithm.
Such kind of learning-based strategies are also known as learning to branch.
Learning to branch is different from the traditional optimization methods.
It introduces the concept of learning in the optimization process to help search the optimal solution more effectively. \cite{Balcan:5} has shown empirically and theoretically that it is possible to learn high-performing branching strategies for a given application domain.
Learning branching policies for MILP has become an active research area.
Most relevant researches use supervised or imitation learning to imitate SB method and specialize it to distinct classes of problems.
However, only the data collected by expert policies can be used to train during the imitation learning process in their studies, which leads to mismatch between training data and real-world data.
This factor prevents learning a good enough policy by the means of imitation learning.
Some other scholars attempt to use reinforcement learning and model the variable selection process as a Markov Decision Process (MDP), to obtain more effective and non-myopic policies.
However, for each instance, the MDP contains a large number of steps and actions, which can lead to a large variance in gradient estimation. Besides, too large action set makes it hard to explore in MILP.
These challenges restrict the applications of reinforcement learning in solving MILP~\cite{sun2020improving}.

In this work, we attempt to address the above challenges by proposing a novel reinforcement learning-based branching algorithm.
Although there have been some ideas to solve the branching problem by means of reinforcement learning~\cite{sun2020improving}, we propose to address the learning problem in a novel way, of which the contributions are summarized as the following points:
\begin{itemize}
    \item Demonstration data is leveraged to accelerate the learning massively at the early stage. With the improvement of the training effect, the agent starts to interact with the environment with its learned policy gradually.
    \item The RL agent updates its network with a mixture of demonstration and self-generated data. We introduce a prioritized storage mechanism to control the mixing ratio automatically.
    \item In order to improve the robustness of the training process, a superior Q-network is additionally introduced based on Double DQN, which always serves as a Q-network with competitive performance.
    \item We evaluate the performance of the proposed algorithm on three benchmarks with comparative experiments against three heuristic approaches and one state-of-the-art (SOTA) imitation learning-based branching algorithm ~\cite{gasse2019exact}. With the dual integral as a metric, our algorithm outperforms the SOTA imitation learning-based branching algorithm by 35.88\% at most.
    Besides, we also make an ablation study to validate the rationality of this algorithm.
\end{itemize}

\section{Related work}

Supervised machine learning and imitation learning are currently the mainstream approaches for learning to branch.
~\cite{Alvarez:6} proposed a new approach that uses supervised learning to improve the performances of optimization algorithms in the context of MILP.
~\cite{khalil2016learning} proposed a machine learning framework for variable branching in MILP. Based on the data collected by SB method, they learned an easy-to-evaluate surrogate function that mimics the SB method, by means of solving a learning-to-rank problem. And it is competitive with a state-of-the-art commercial solver.
~\cite{gasse2019exact} proposed a new graph convolutional neural network (GCNN) model for learning to branch, which leverages the natural variable-constraint bipartite graph representation of MILP. They trained the GCNN model via imitation learning from the SB method, and demonstrated that this model produced policies that improved upon state-of-the-art machine learning methods for branching.
~\cite{gupta2020hybrid} proposed a new hybrid architecture for efficient branching on CPU machines, which combined the expressive power of GCNNs with computationally inexpensive multi-layer perceptrons (MLPs) for branching.

In order to obtain more efficient and non-myopic policies, some scholars attempt to use reinforcement learning to solve this problem.
~\cite{etheve2020reinforcement} proposed Fitting for Minimising the SubTree Size, a novel approach based on reinforcement learning, whose strength lies in the consistency between a local value function and a global metric of interest.
~\cite{sun2020improving} introduced a novel set representation and optimal transport distance for the branching process associated with a policy, to train the reinforcement learning agent. The results showed substantial improvements in empirical evaluation.
More related researches can refer to the survey provided by ~\cite{huang2021branch}.
\section{Background}

\subsection{Mixed integer linear programs}

A mixed integer linear program is an optimization problem of the form
\[ \mathop{\arg\min}\limits_{\textbf{x}} \left\{ \textbf{c}^T\textbf{x} \mid \textbf{Ax} \leq \textbf{b}, \textbf{l} \leq \textbf{x} \leq \textbf{u}, \textbf{x} \in \mathbb{Z}^\textit{p} \times \mathbb{R}^\textit{n-p} \right\} \]
where $\textbf{c} \in \mathbb{R}^\textit{n}$ is the objective coefficient vector, 
$\textbf{A} \in \mathbb{R}^{\textit{m} \times \textit{n}}$ is the constraint coefficient matrix, 
$\textbf{b} \in \mathbb{R}^\textit{m}$ is the constraint right-hand-side vector, 
$\textbf{l}, \textbf{u} \in \mathbb{R}^\textit{n}$ represent the lower and upper variable bound vectors respectively, 
and $\textit{p}$ is the number of integer variables.
The linear programming (LP) relaxation of a MILP is shown below
\[ \mathop{\arg\min}\limits_{\textbf{x}} \left\{ \textbf{c}^T\textbf{x} \mid \textbf{Ax} \leq \textbf{b}, \textbf{l} \leq \textbf{x} \leq \textbf{u}, \textbf{x} \in \mathbb{R}^\textit{n} \right\} \]
The LP solution provides a lower bound to the original MILP. Specifically, if the LP solution is subject to the integer constraint, then it is also a optimal feasible solution of the MILP. Otherwise, the LP relaxation is required to be decomposed into two sub-problems.
This is done by branching on a variable that does not obey the integrality constraint in the current LP solution.
The solving process terminates when the feasible regions cannot be decomposed anymore, and subsequently a certificate of optimality or infeasibility can be provided respectively.

\subsection{Branch-and-bound}

A key factor influencing the efficiency of B\&B algorithm is how to select a fractional variable to branch on.
There are commonly two basic heuristic branching rules, including Strong Branching (SB) and Pseudocost Branching (PC) strategies. Most other heuristic rules are derived from them.
The details of SB and PC are described as follows.

SB is to evaluate which of the fractional candidates gives the best progress before actually branching on any of them.
For each candidate variable, this evaluation process is realized by solving the LP relaxations of the two sub-problems.
Thus, a huge amount of computation is required when adopting SB method.

PC is a sophisticated rule in the sense that it keeps a history of the success of the variable on which already has been branched.
$\Psi_j^+$ ($\Psi_j^-$) denotes the average unit objective gain taken over upwards (downwards) branching on $x_j$ in previous nodes.
Pseudocost branching at node $\textit{N}$ with LP relaxation solution $\check{x}$ consists in computing values:
\[ PC_j = score((\check{x}_j - \left \lfloor \check{x}_j \right \rfloor)\Psi_j^-,(\left \lceil \check{x}_j \right \rceil - \check{x}_j)\Psi_j^+) \]
and choosing the candidate variable with highest such value.
Compared with the SB method, PC method is simpler but faster.
However, PC method is overly dependent on human intuition and extensive engineering, requiring considerable manual tuning.

\subsection{Reinforcement learning for branching}

According to the reference~\cite{gasse2019exact}, the B\&B process can be modeled as a Markov decision process (MDP), in which the solver is considered as the environment.
Prouvost et al. presented Ecole, a new library to simplify machine learning research for combinatorial optimization, which lowers the bar of entry and accelerates innovation in this field~\cite{prouvost2020ecole}.

Based on Ecole, the B\&B is episodic. Each episode corresponds to a MILP instance. The state and reward of the MDP can be defined by an observation function and a reward function in Ecole respectively.
Or even users can define new environments and simply reuse existing observation and reward functions to fulfill specific requirements.
The probability of a trajectory $ \tau = (s_0, ..., s_T) \in T $ depends on both the branching policy $\pi$ and the remaining components of the solver,
\[ p_\pi(\tau) = p(s_0) \prod_{t=0}^{T-1} \sum_{a \in A(s_t)} \pi(a \mid s_t) p(s_{t+1} \mid s_t, a) \]

Compared with the imitation learning, the reinforcement learning can balance the exploration and exploitation in an unknown environment. Theoretically speaking, it possesses higher performance.

\section{Proposed algorithm}

In this paper, a novel algorithm based on reinforcement learning is proposed for dealing with the B\&B variable selection problem in MILP.
In order to illustrate our algorithm, we organize this section in two parts.
Firstly, we give our formulation of the variable selection process as a reinforcement learning problem.
Then, we introduce the specific implementation of this algorithm in detail.

\subsection{Formulation}

Let the solver be the environment. The sequential decision making of variable selection can be formulated as a Markov Decision Process (MDP).
Specifically, the state \textit{s}, action space \textit{A}, action \textit{a}, transition \textit{P}, and reward \textit{r} are described as follows.

\begin{itemize}
    \item State: A node bipartite graph representation of B\&B states used in Gasse et al.~\cite{gasse2019exact}, using the ecole.observation.NodeBipartite observation function~\cite{prouvost2020ecole}. 
    \item Action space: The set of candidate variables.
    \item Action: The selected candidate variable to branch on.
    \item Transition: Given state $ s_t $ and action $ a_t $, the next state $ s_{t+1} $ is determined by the node selection policy.
    \item Reward: The reward is defined as the dual integral since the previous state, where the integral is computed with respect to the solving time. Details are explained in the Metric part of the Experimental evaluation section.
    \item Next state: The node bipartite graph representation of the next node.
    \item Next action space: The set of candidate variables corresponding to the next node.
    \item Done: The termination flag.
\end{itemize}

\subsection{Algorithm}

In this section, based on Double DQN~\cite{van2016deep}, we propose a novel deep reinforcement learning algorithm for branching problem.
It leverages demonstration data to massively accelerate the learning process by means of offline reinforcement learning.
In order to balance the exploration and exploitation, this algorithm can automatically control the mixing ratio of demonstration data during the learning process due to a prioritized storage mechanism.
Besides, in order to avoid the multi-fold challenges caused by the large state space and action set, we additionally introduce a superior network, which always serves as a Q-network with competitive performance.
Figure 1 shows an overview of the proposed algorithm.
\begin{figure*}[htp]
    \centering
    \includegraphics[height=6.8cm]{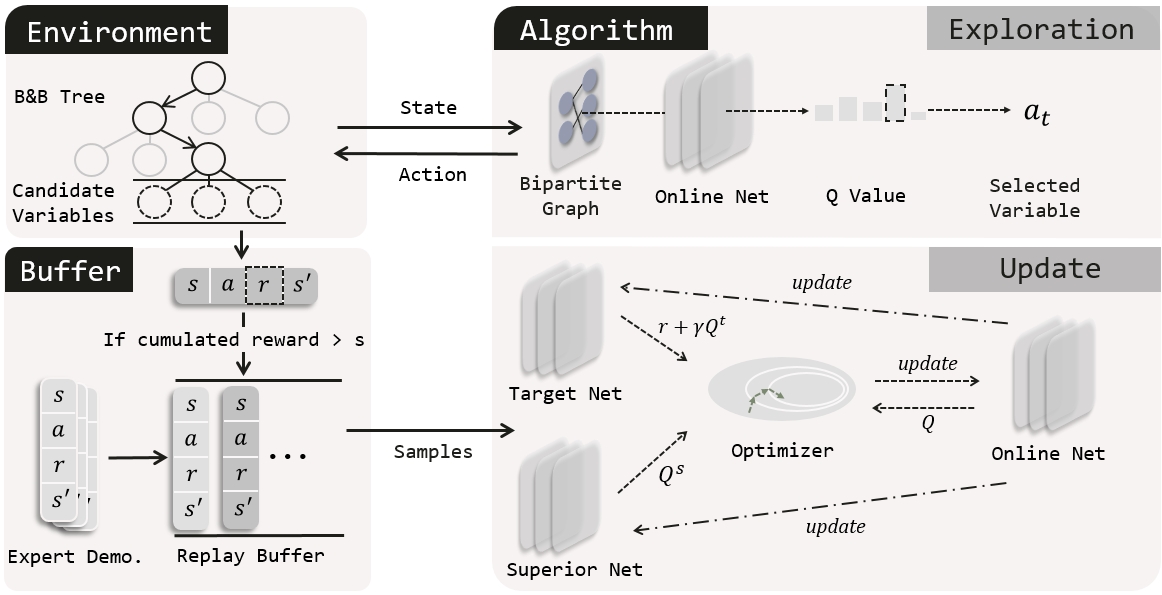}
    \caption{An overview of the proposed algorithm: Environment corresponds to a solver aiming at variable selection problem in B\&B; Buffer is filled with both demonstration and self-generated data, and only high-quality self-generated data can be added in it; The main part of the algorithm is composed of three networks, where online net is used for action selection, target net is used for action evaluation to decompose the max operation, superior net always serves as a Q-network with competitive performance to help to stabilize the training process.}
    \label{fig:overview}
\end{figure*}

In order to accelerate the learning process at the early stage, the replay buffer is given a set of demonstration data, which will remain permanently.
In this phase, the agent trains solely on the demonstration data by means of offline reinforcement learning, without any interaction with the environment.
And it can obtain relatively good performance in a short period of time.
However, the agent may make incorrect estimates for unfamiliar $ (s,a) $ pairs due to the lack of interaction with the environment.
Therefore, we introduce a prioritized storage mechanism, in which the agent collects self-generated data and adds it to the replay buffer conditionally.
In detail, only the data collected by the policy that leads to high performance is added to the replay buffer, which can prevent some poor-quality data from misleading the training.
Data is added to the buffer until it is full, and then agent starts overwriting old data in the buffer.
However, the agent never overwrites the demonstration data.
In this way, the mixing ratio between demonstration and self-generated data can be controlled automatically.


It is commonly believed that too large action set increases the difficulty of RL training~\cite{sun2020improving}.
The RL agent may learn a local optimal policy, or even worse, a random policy.
In order to solve this issue, a superior Q-network is additionally introduced based on the framework of Double DQN.
And then, there are three Q-networks in this frame, i.e. online network, target network, and superior network.
The online network is used for action selection. The target network is used for action evaluation to decompose the max operation. The superior network always serves as a Q-network with competitive performance to help to stabilize the training process.
Similar to the works of Gasse et al., a graph convolutional neural network (GCNN) is adopted to parametrize the Q-network~\cite{gasse2019exact}. 
In detail, the input of the GCNN model is the bipartite state representation $s_t = (\mathcal{G}, \textbf{C}, \textbf{V}, \textbf{E})$.
The graph convolution can be broken down into two successive passes, one from variables to constraints and the other one from constraints to variables, which are shown as
\begin{align*}
    c_i &\gets f_c(c_i, \sum_j^{(i,j) \in \varepsilon} {g_c(c_i,v_j,e_{i,j})}) \\
    v_i &\gets f_v(v_j, \sum_j^{(i,j) \in \varepsilon} {g_v(c_i,v_j,e_{i,j})})
\end{align*}
where $f_c$, $f_v$, $g_c$ and $g_v$ are two-layer perceptrons with ReLU activation functions.
In addition to TD loss used in Double DQN, we add another term in the overall loss used to update the online network, which is shown as
\begin{align*}
    \mathcal{L}(\theta) = &\mathbb{E}(r + \gamma \max_{a^{\prime}} Q^t(s^{\prime},a^{\prime},\theta^t) - Q(s,a,\theta)) \\ + &\mathbb{E}(Q^s(s,a,\theta^s) - Q(s,a,\theta))
\end{align*}
where the superscript \textit{t} and \textit{s} represent \textit{target} and \textit{superior} respectively.

Pseudo-code is sketched in Algorithm 1. The behavior policy $ \pi^{\varepsilon Q_\theta} $ is $\varepsilon$-greedy with respect to $ Q_\theta $.

\begin{algorithm}[t]
	\caption{Double Deep Q-learning with Superior Network} 
	\label{alg} 
	\begin{algorithmic}[1]
		\REQUIRE $ D^{replay} $: initialized with demonstration data set; \\\ $ \theta $: weights for online network (random initialization); \\\ $ \theta^t $: weights for target network (random initialization);
		\\\ $ \theta^s $: weights for superior network (random initialization); \\\ $ \tau^t $: frequency at which to update target network; \\\ $ \tau^s $: frequency at which to evaluate the trained policy; \\\ $ G_0 $: a threshold to measure the performance of behavior policy; \\\ $ G_{best} $: the cumulative reward of the best trained policy in the training process (zero initialization).
		\FOR{steps $ t \in {1,2,...} $}
		\STATE Sample action from behavior policy $ a \sim \pi^{\varepsilon Q_\theta} $
		\STATE Play action \textit{a} and observe $ (s^\prime, r) $
		\STATE Store $ (s,a,r,s^\prime) $ into a temporary buffer $ D^{temporary} $
		\STATE Sample a mini-batch of \textit{n} transitions from $ D^{replay} $
		\STATE Calculate loss $ \mathcal{L} $ using target and superior network
		\STATE Perform a gradient descent step to update $ \theta $
		\IF{$ t \mod \tau^t = 0 $}
		\STATE $ \theta^t \gets \theta $
		\ENDIF
		\IF{$ t \mod \tau^s = 0 $}
		\STATE Compute the cumulative reward \textit{G} to evaluate the trained policy
		\IF{$ G > G_0 $}
		\STATE Store $ (s,a,r,s^\prime) $ in $ D^{temporary} $ into $ D^{replay} $, overwriting oldest self-generated transition if over capacity, emptying $ D^{temporary} $
		\ENDIF
		\IF{$ G > G_{best} $}
		\STATE $ \theta^s \gets \theta $
		\ENDIF
		\ENDIF
		\STATE $ s \gets s^\prime $
		\ENDFOR
	\end{algorithmic} 
\end{algorithm}

\section{Experimental evaluation}

In this part, we achieve a comparative experiment against three heuristic approaches and one machine learning approach to evaluate the effectiveness of our algorithm.
Meanwhile, we also make an ablation study to validate the rationality of this algorithm.

\subsection{Setup}

The proposed method is trained on a computing server which is equipped with Intel(R) Xeon(R) Platinum 8180M CPU@2.50GHz, a V100 GPU card with 32GB graphic memory and 1TB main memory.

\textbf{Benchmark}:
We evaluate our algorithm on three benchmarks from diverse application areas, including \textit{Balanced Item Placement} (\textit{BIP}), \textit{Workload Apportionment}  (\textit{WA}), and an \textit{Anonymous Problem}  (\textit{AP}). All the three benchmarks are from the Machine Learning for Combinatorial Optimization (ML4CO) NeurIPS 2021 competition~\cite{ml4co}. The first two benchmarks are inspired by real-life applications of large-scale systems at Google, and the third benchmark is an anonymous problem also inspired by a real-world, large-scale industrial application.

\textbf{Metric}:
The dual integral is used as the evaluation metric, which measures the area over the curve of the solver's dual bound (a.k.a. global lower bound).
It usually corresponds to a solution of a valid relaxation of the MILP.
Theoretically, a small dual integral will lead to both good and fast decisions.
By branching, the LP relaxations corresponding to the branch-and-bound tree leaves get tightened, and the dual bound increases over time. 
With a time limit \textit{T}, the dual integral expresses as:
\[ T c^T x^* - \int_{t=0}^T z_t^* d t \]
where $ z_t^* $ is the best dual bound at time \textit{t}, and $ T c^T x^* $ is an instance-specific constant that depends on the optimal solution value $ c^T x^* $.
The dual integral is to be minimized, and takes an optimal value of 0.
The optimization process is shown in Figure 2.
In the experiments of this paper,  the time limit is set as 15 minutes.
\begin{figure}[htp]
    \centering
    \includegraphics[width=8cm]{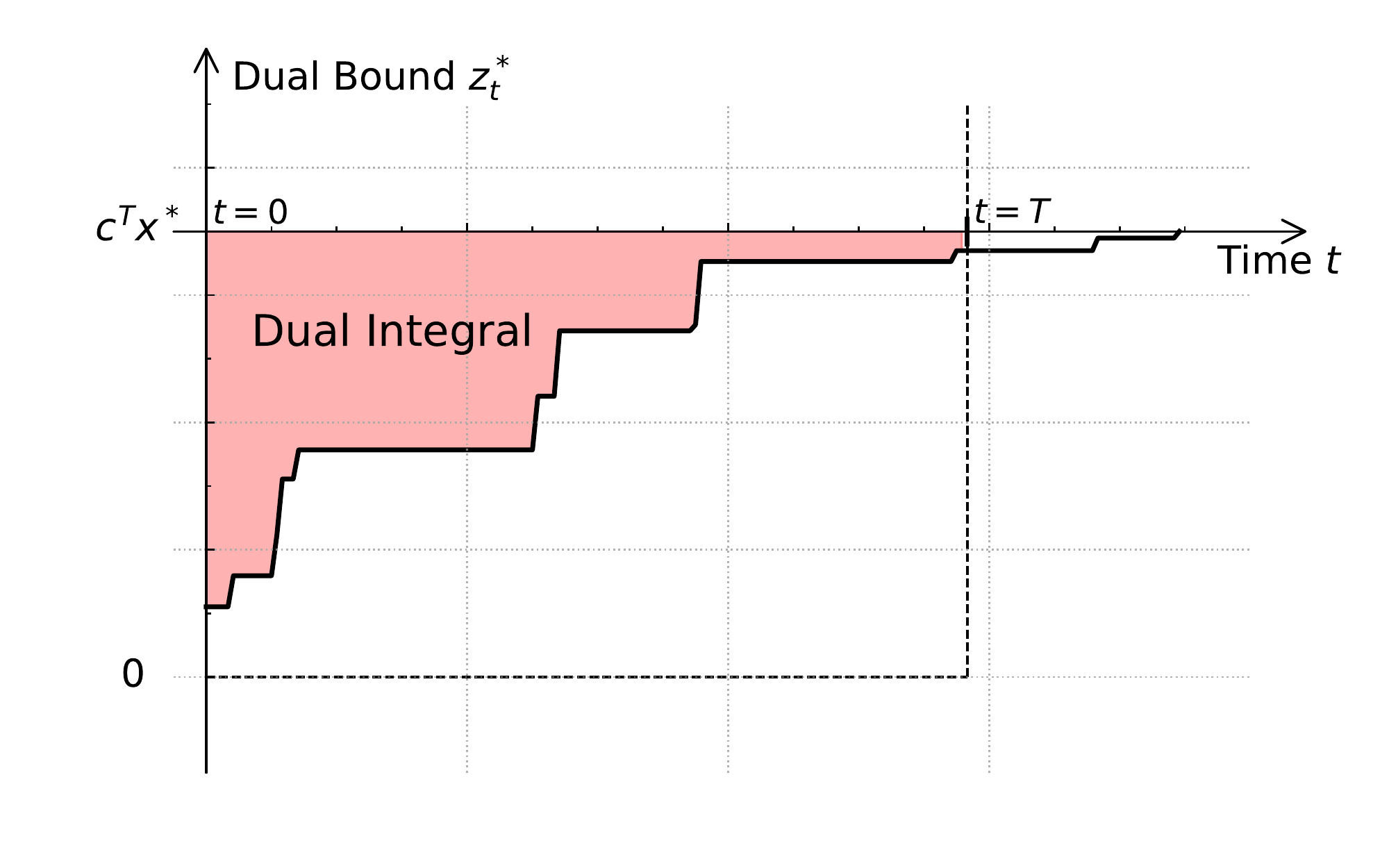}
    \caption{The dual integral during the optimization process}
    \label{fig:dual integral}
\end{figure}

\textbf{Baselines}:
We compete against three heuristic approaches, including strong branching, pseudocost branching, active constraint method (a method suitable for solving large-scale problems)~\cite{patel2007active}, and one machine learning approach, i.e. imitation learning from the strong branching expert rule~\cite{gasse2019exact}.

\subsection{Comparative experiment}

In this experiment, the demonstration data set size is 50K for \textit{BIP} and \textit{AP} datasets, and 10K for the \textit{WA} dataset.

\begin{itemize}
    \item Batch size: 32 for \textit{BIP} and \textit{AP}, 24 for \textit{WA}
    \item Learning rate: 0.001
    \item $ \varepsilon $-greedy with $ \varepsilon = 0.01 $
    \item Dimension of input embedding in GCNN: 64
    \item Replay buffer size: 100K for \textit{BIP} and \textit{AP}, 20K for \textit{WA}
    \item Discount factor: 0.99
    \item Period for target model's hard update: $ \tau^t $ = 500
    \item Period for evaluating the trained policy: $ \tau^s $ = 1000
    \item Epochs: 50K
\end{itemize}

\begin{table}[!h]
\centering
\begin{tabular}{lll}
\hline
\makecell[c]{Algorithms}  & \makecell[c]{Scores} & \makecell[c]{Wins} \\
\hline
\makecell[c]{SB} & \makecell[c]{3767.95} & \makecell[c]{0/100} \\
\makecell[c]{PC} & \makecell[c]{4268.95} & \makecell[c]{1/100} \\
\makecell[c]{AC} & \makecell[c]{5800.05} & \makecell[c]{3/100} \\
\makecell[c]{IL} & \makecell[c]{5335.94} & \makecell[c]{33/100} \\
\makecell[c]{RL} & \makecell[c]{\textbf{7250.29}} & \makecell[c]{\textbf{63/100}} \\
\hline
\end{tabular}
\caption{Evaluation results on \textit{Balanced Item Placement}}
\label{tab:plain}
\end{table}
\begin{table}[!h]
\centering
\begin{tabular}{lll}
\hline
\makecell[c]{Algorithms}  & \makecell[c]{Scores} & \makecell[c]{Wins} \\
\hline
\makecell[c]{SB} & \makecell[c]{623469.3} & \makecell[c]{13/100} \\
\makecell[c]{PC} & \makecell[c]{624727.1} & \makecell[c]{18/100} \\
\makecell[c]{AC} & \makecell[c]{624256.8} & \makecell[c]{14/100} \\
\makecell[c]{IL} & \makecell[c]{624326.4} & \makecell[c]{20/100} \\
\makecell[c]{RL} & \makecell[c]{\textbf{624932.8}} & \makecell[c]{\textbf{35/100}} \\
\hline
\end{tabular}
\caption{Evaluation results on \textit{Workload Apportionment}}
\label{tab:plain}
\end{table}
\begin{table}[!h]
\centering
\begin{tabular}{lll}
\hline
\makecell[c]{Algorithms}  & \makecell[c]{Scores} & \makecell[c]{Wins} \\
\hline
\makecell[c]{SB} & \makecell[c]{30387450} & \makecell[c]{27/100} \\
\makecell[c]{PC} & \makecell[c]{31600684} & \makecell[c]{1/100} \\
\makecell[c]{AC} & \makecell[c]{30995284} & \makecell[c]{1/100} \\
\makecell[c]{IL} & \makecell[c]{32446178} & \makecell[c]{29/100} \\
\makecell[c]{RL} & \makecell[c]{\textbf{32547931}} & \makecell[c]{\textbf{42/100}} \\
\hline
\end{tabular}
\caption{Evaluation results on \textit{Anonymous Problem}}
\label{tab:plain}
\end{table}

We evaluate different branching algorithms on 100 instances for each benchmarks.
Evaluation results on three benchmarks are shown in Tables 1-3 respectively.

\begin{figure*}[htp]
    \centering
    \subfigure[\textit{Balanced Item Placement}]{
    \includegraphics[width=5.5cm]{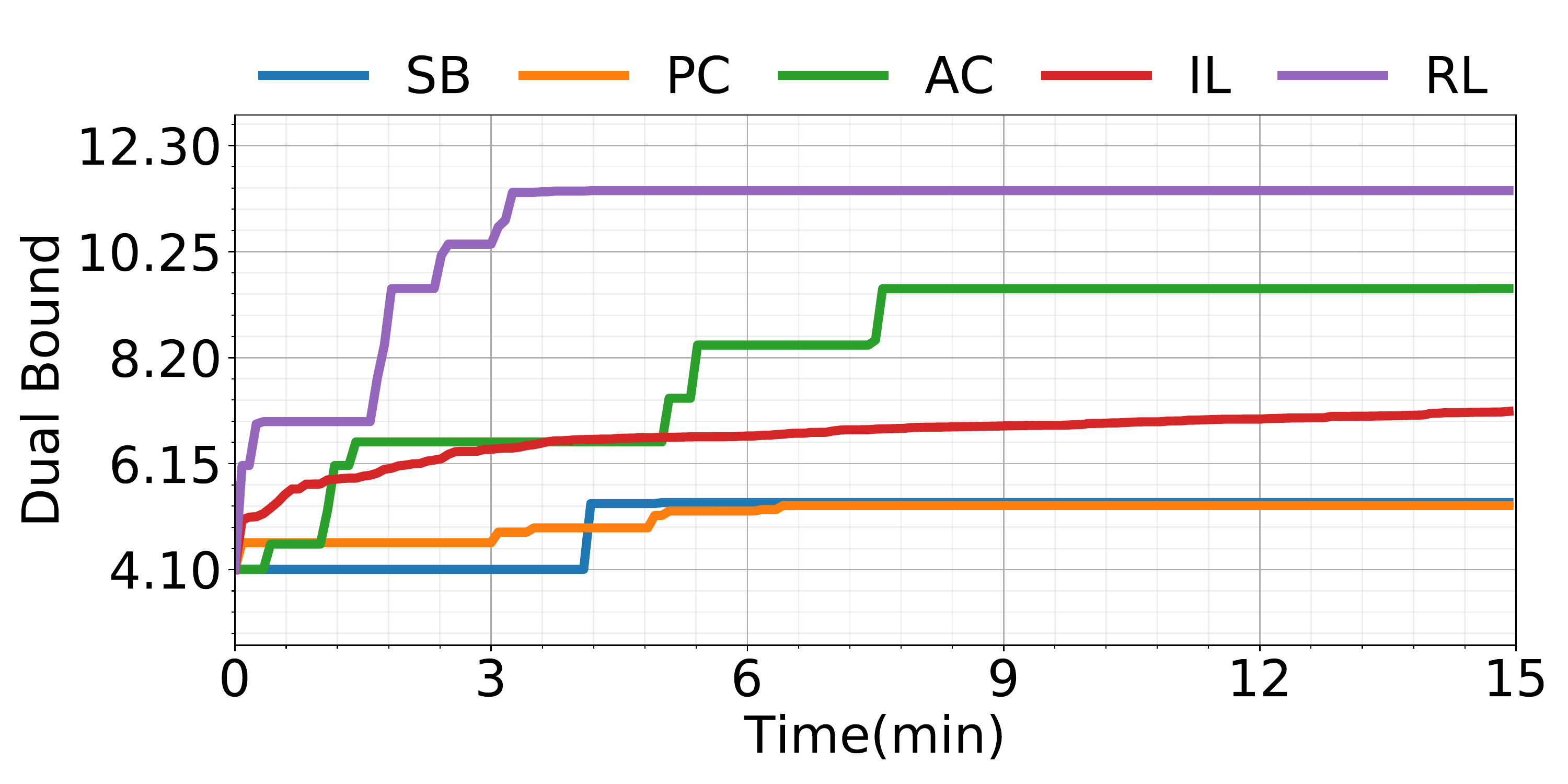}
    }
    \subfigure[\textit{Workload Apportionment}]{
    \includegraphics[width=5.5cm]{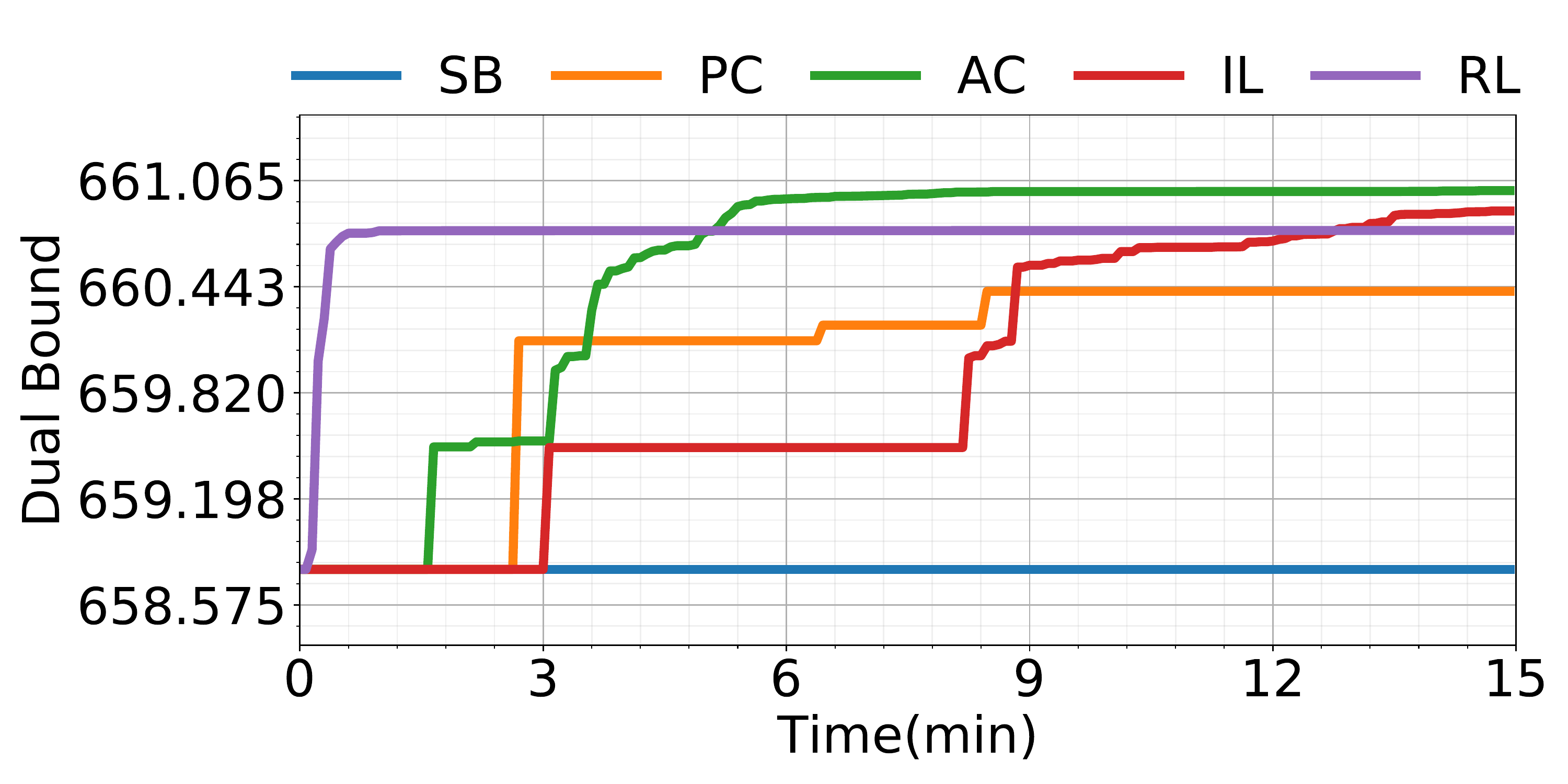}
    }
    \subfigure[\textit{Anonymous Problem}]{
    \includegraphics[width=5.5cm]{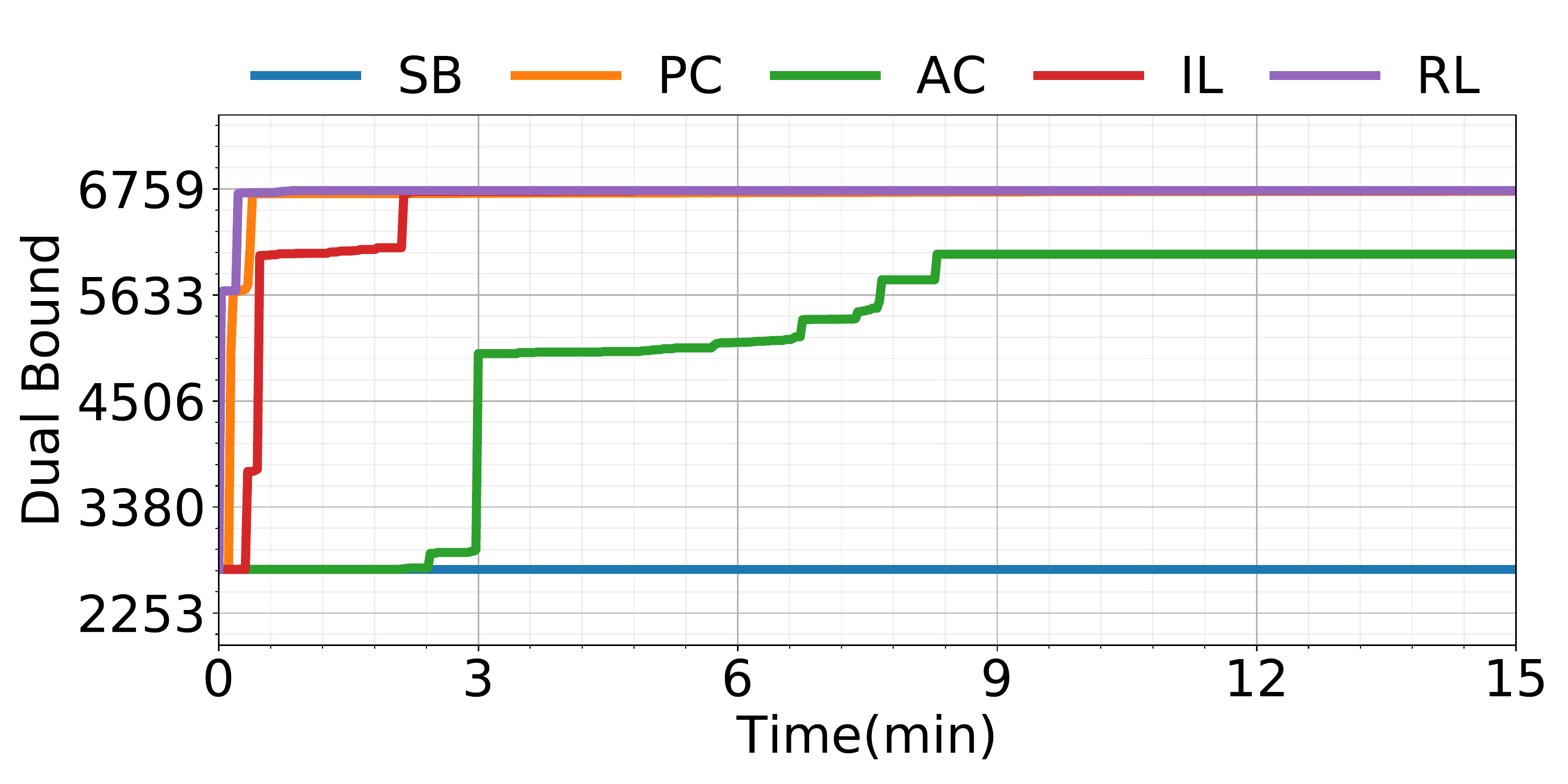}
    }
    \caption{The dual bound curve of different algorithms during the solving process}
    \label{fig:training process}
\end{figure*}
\begin{figure*}[htp]
    \centering
    \subfigure[\textit{Balanced Item Placement}]{
    \includegraphics[width=5.5cm]{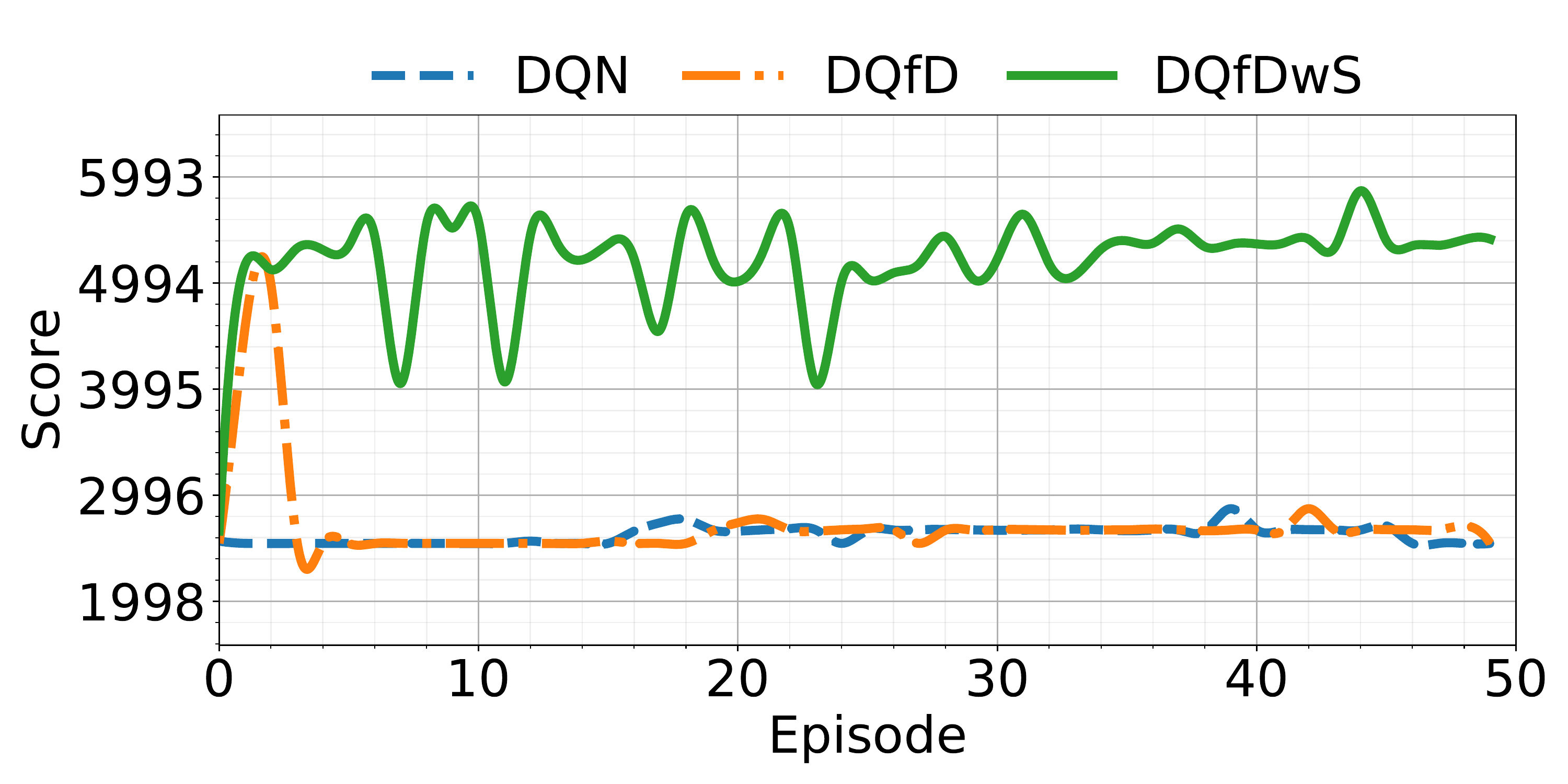}
    }
    \subfigure[\textit{Workload Apportionment}]{
    \includegraphics[width=5.5cm]{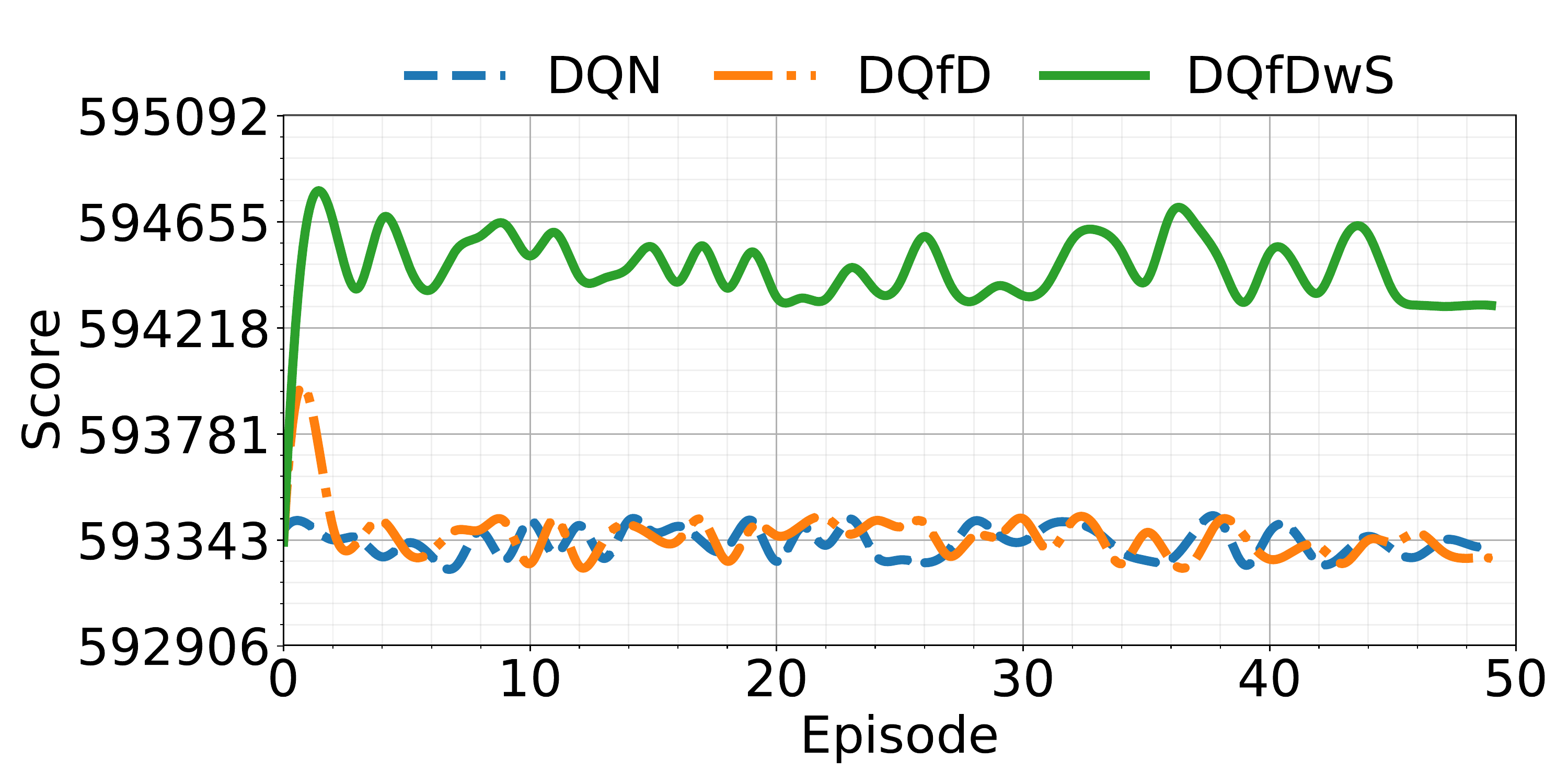}
    }
    \subfigure[\textit{Anonymous Problem}]{
    \includegraphics[width=5.5cm]{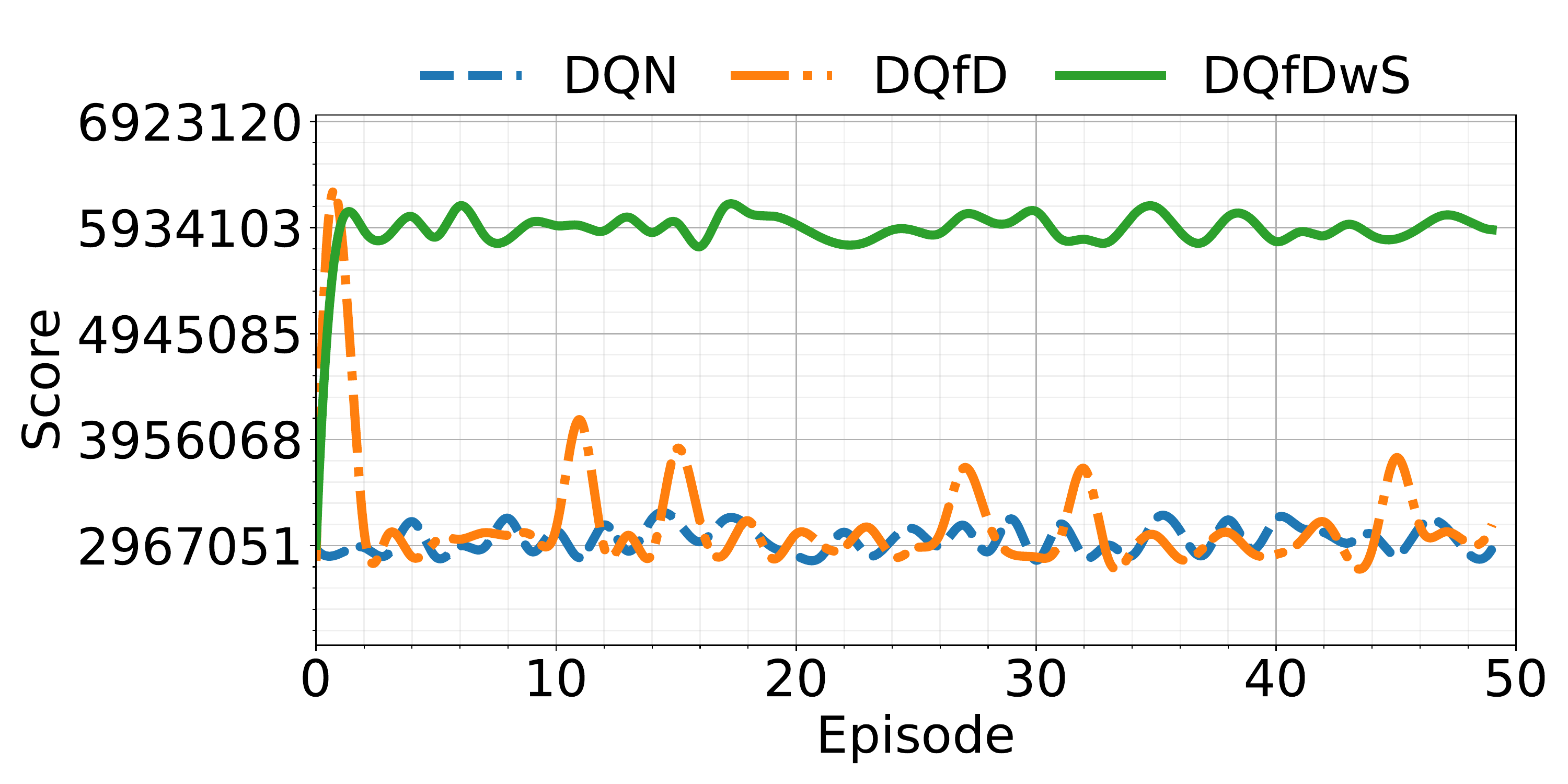}
    }
    \caption{The performance curve of ablation study in the training process}
    \label{fig:training process}
\end{figure*}

In Tables 1-3, SB, PC, AC, IL, RL represent the strong branching, pseudocost branching, active constraint method, imitation learning from the strong branching expert rule, and our algorithm respectively.

As can be seen, AC is indeed more suitable for solving large-scale MILP problems among the three heuristic algorithms. However, its performance is still inferior to our algorithm.
Furthermore, it is worth mentioning that both the imitation learning algorithm proposed by Gasse et al. and our reinforcement learning algorithm learn from the demonstration data collected by SB and PC approaches.
It can be observed that both machine learning algorithms have a greater performance improvement than expert rules.
Meanwhile, under the same conditions including the data set size, learning rate, etc, the performance of our algorithm is significantly better than that of the imitation learning algorithm, which is demonstrated by the facts that our algorithm obtains the highest scores and wins on the largest number of instances for all three benchmarks.
Compared with IL, this proposed algorithm possesses a performance improvement of up to 35.88\%, exactly on the first benchmark.

In order to further illustrate the effect of different algorithms, we perform branch and bound on one instance for each benchmark, and give the dual bound curve in this process, which are shown in Figure 3.
It can be seen that our algorithm can indeed result in the smallest dual integral, which suggests that it can lead to both the best and fastest decisions.

\vspace{-1ex}
\subsection{Ablation study}

We present an ablation study of our proposed algorithm by comparing with two reinforcement learning algorithms, including the deep reinforcement learning with double Q-learning and the double deep Q-learning only with demonstration data.
We plot the performance curves of different algorithms on test dataset during training process for all three benchmarks.
The results are shown in Figure 4. DQN, DQfD and DQfDwS represent the Double DQN, Double DQN with the guidance of expert rules and our algorithm respectively.
It can be found that the performance of deep reinforcement learning with double Q-learning is extremely poor to solve the MILP problems, in which the trained policy are almost always random.
After adding the guidance of expert rules, it can be seen that the training process is greatly accelerated in the early stage of training, but it will soon fall into the local optimal strategy, which is caused by the too large state space and action set.
However, after adding the superior Q-network, our algorithm can not only help to accelerate the training process in the early stage but also help to stabilize the training process in the late stage.
Thus our algorithm has the potential to advance the performance of B\&B algorithm continuously.
\vspace{-4ex}
\vspace{-1ex}

\section{Conclusion}

In this paper, we propose a novel reinforcement learning method to improve the B\&B algorithm performance in MILP.
Firstly, in view of the shortcomings of reinforcement learning algorithms that are difficult to explore in the early stage of this problem, demonstration data collected by strong branch rule is leveraged to accelerate the learning process significantly.
Subsequently, as the training effect improves, the agent gradually begins to interact with the solver with its learned policy.
To avoid misleading training by low-quality self-generated data, a prioritized storage mechanism is introduced to guarantee the high quality of data in the replay buffer and control the mixing ratio of demonstration and self-generated data automatically.
Besides, too large state space and action set lead to a large variance in gradient estimation, which makes it hard to explore in MILP. In order to address this challenge, we additionally introduced a superior Q-network based on Double DQN, which always serves as a Q-network with competitive performance.
A group of comparative experiments and ablation studies demonstrate that this algorithm is significantly effective in performance improvement of B\&B algorithm.

\clearpage
\bibliographystyle{named}
\bibliography{ijcai22}

\begin{thebibliography}{}

\bibitem[\protect\citeauthoryear{Achterberg \bgroup \em et al.\egroup
  }{2005}]{Achterberg:2}
Tobias Achterberg, Thorsten Koch, and Alexander Martin.
\newblock Branching rules revisited.
\newblock {\em Operations Research Letters}, 33(1):42--54, Jan 2005.

\bibitem[\protect\citeauthoryear{Alvarez \bgroup \em et al.\egroup
  }{2014a}]{Alvarez14asupervised}
Alejandro~Marcos Alvarez, Quentin Louveaux, and Louis Wehenkel.
\newblock A supervised machine learning approach to variable branching in
  branch-and-bound.
\newblock In {\em IN ECML}, 2014.

\bibitem[\protect\citeauthoryear{Alvarez \bgroup \em et al.\egroup
  }{2014b}]{Alvarez:6}
Marcos Alvarez, Alejandro, Quentin Louveaux, and Louis Wehenkel.
\newblock A supervised machine learning approach to variable branching in
  branch-and-bound.
\newblock 2014.

\bibitem[\protect\citeauthoryear{Applegate \bgroup \em et al.\egroup
  }{1995}]{Applegate:4}
D~Applegate, R~Bixby, V~Chvátal, and W~Cook.
\newblock Finding cuts in the tsp.
\newblock {\em DIMACS Technical Report 95-05}, Mar 1995.

\bibitem[\protect\citeauthoryear{Balcan \bgroup \em et al.\egroup
  }{2018}]{Balcan:5}
MF~Balcan, T~Dick, T~Sandholm, and E~Vitercik.
\newblock Learning to branch.
\newblock In {\em Proceedings of the 35th International Conference on Machine
  Learning}, Stockholm, Sweden, July 2018. ACM.

\bibitem[\protect\citeauthoryear{Benichou \bgroup \em et al.\egroup
  }{1971}]{Benichou:3}
M.~Benichou, J.~M. Gauthier, P.~Girodet, G.~Hentges, G.~Ribiere, and
  O.~Vincent.
\newblock Experiments in mixed-integer linear programming.
\newblock {\em Mathematical Programming}, 1:76--94, Dec 1971.

\bibitem[\protect\citeauthoryear{Etheve \bgroup \em et al.\egroup
  }{2020}]{etheve2020reinforcement}
Marc Etheve, Zacharie Al{\`e}s, C{\^o}me Bissuel, Olivier Juan, and Safia
  Kedad-Sidhoum.
\newblock Reinforcement learning for variable selection in a branch and bound
  algorithm.
\newblock In {\em International Conference on Integration of Constraint
  Programming, Artificial Intelligence, and Operations Research}, pages
  176--185. Springer, 2020.

\bibitem[\protect\citeauthoryear{Gasse \bgroup \em et al.\egroup
  }{2019}]{gasse2019exact}
Maxime Gasse, Didier Ch{\'e}telat, Nicola Ferroni, Laurent Charlin, and Andrea
  Lodi.
\newblock Exact combinatorial optimization with graph convolutional neural
  networks.
\newblock {\em arXiv preprint arXiv:1906.01629}, 2019.

\bibitem[\protect\citeauthoryear{Gupta \bgroup \em et al.\egroup
  }{2020}]{gupta2020hybrid}
Prateek Gupta, Maxime Gasse, Elias~B Khalil, M~Pawan Kumar, Andrea Lodi, and
  Yoshua Bengio.
\newblock Hybrid models for learning to branch.
\newblock {\em arXiv preprint arXiv:2006.15212}, 2020.

\bibitem[\protect\citeauthoryear{Huang \bgroup \em et al.\egroup
  }{2021}]{huang2021branch}
Lingying Huang, Xiaomeng Chen, Wei Huo, Jiazheng Wang, Fan Zhang, Bo~Bai, and
  Ling Shi.
\newblock Branch and bound in mixed integer linear programming problems: A
  survey of techniques and trends.
\newblock {\em arXiv preprint arXiv:2111.06257}, 2021.

\bibitem[\protect\citeauthoryear{Khalil \bgroup \em et al.\egroup
  }{2016}]{khalil2016learning}
Elias Khalil, Pierre Le~Bodic, Le~Song, George Nemhauser, and Bistra Dilkina.
\newblock Learning to branch in mixed integer programming.
\newblock In {\em Proceedings of the AAAI Conference on Artificial
  Intelligence}, volume~30, 2016.

\bibitem[\protect\citeauthoryear{{NeurIPS 2021 Competition}}{2021}]{ml4co}
{NeurIPS 2021 Competition}.
\newblock Machine learning for combinatorial optimization,
  {\url{https://www.ecole.ai/2021/ml4co-competition/}},, 2021.

\bibitem[\protect\citeauthoryear{Patel and Chinneck}{2007}]{patel2007active}
Jagat Patel and John~W Chinneck.
\newblock Active-constraint variable ordering for faster feasibility of mixed
  integer linear programs.
\newblock {\em Mathematical Programming}, 110(3):445--474, 2007.

\bibitem[\protect\citeauthoryear{Prouvost \bgroup \em et al.\egroup
  }{2020}]{prouvost2020ecole}
Antoine Prouvost, Justin Dumouchelle, Lara Scavuzzo, Maxime Gasse, Didier
  Ch{\'e}telat, and Andrea Lodi.
\newblock Ecole: A gym-like library for machine learning in combinatorial
  optimization solvers.
\newblock {\em arXiv preprint arXiv:2011.06069}, 2020.

\bibitem[\protect\citeauthoryear{Sun \bgroup \em et al.\egroup
  }{2020}]{sun2020improving}
Haoran Sun, Wenbo Chen, Hui Li, and Le~Song.
\newblock Improving learning to branch via reinforcement learning.
\newblock 2020.

\bibitem[\protect\citeauthoryear{Van~Hasselt \bgroup \em et al.\egroup
  }{2016}]{van2016deep}
Hado Van~Hasselt, Arthur Guez, and David Silver.
\newblock Deep reinforcement learning with double q-learning.
\newblock In {\em Proceedings of the AAAI conference on artificial
  intelligence}, volume~30, 2016.

\end{thebibliography}


\end{document}